\pdfoutput=1

\documentclass[11pt]{article}

\usepackage{acl}

\usepackage{times}
\usepackage{latexsym}

\usepackage[T1]{fontenc}

\usepackage[utf8]{inputenc}

\usepackage{microtype}

\usepackage{soul}
\usepackage{url}
\usepackage{hyperref}
\usepackage{graphicx}
\usepackage{amsmath}
\usepackage{booktabs}
\usepackage{multirow}
\usepackage{amssymb}    
\usepackage{algorithm}
\usepackage{algorithmic}
\usepackage{arydshln}   
\usepackage{subfig}
\usepackage{ascii}

\title{STAD: Self-Training with Ambiguous Data for Low-Resource Relation Extraction}

\author{
Junjie Yu$^1$ \quad Xing Wang$^{2}$ \quad Jiangjiang Zhao$^{3}$ \quad Chunjie Yang$^{3}$ \quad Wenliang Chen$^1$\\
$^1$Institute of Artificial Intelligence, School of Computer Science and Technology, \\Soochow University, China \\
$^2$Tencent AI Lab, Shenzhen, China \\  
$^3$China Mobile Online Marketing and Services Center, China \\
{\asciifamily \normalsize  jjyu@stu.suda.edu.cn, brightxwang@tencent.com, wlchen@suda.edu.cn} \\ 
{\asciifamily \normalsize \{zhaojiangjiang, yangchunjie\}@cmos.chinamobile.com}
}

\begin{document}

\maketitle

\begin{abstract}
We present a simple yet effective self-training approach, named as STAD, for low-resource relation extraction. The approach first classifies the auto-annotated instances into two groups: confident instances and uncertain instances, according to the probabilities predicted by a teacher model. In contrast to most previous studies, which mainly only use the confident instances for self-training, we make use of the uncertain instances. To this end, we propose a method to identify ambiguous but useful instances from the uncertain instances and then divide the relations into candidate-label set and negative-label set for each ambiguous instance. Next, we propose a set-negative training method on the negative-label sets for the ambiguous instances and a positive training method for the confident instances. Finally, a joint-training method is proposed to build the final relation extraction system on all data. Experimental results on two widely used datasets SemEval2010 Task-8 and Re-TACRED with low-resource settings demonstrate that this new self-training approach indeed achieves significant and consistent improvements when comparing to several competitive self-training systems.\footnote{\ Code is publicly available at \url{https://github.com/jjyunlp/STAD}}
\end{abstract}

\section{Introduction}
\label{intro}
Relation Extraction (RE) is a fundamental task in Information Extraction, which aims to obtain a predefined semantic relation between two entities in a given sentence~\citep{zhou2005exploring}.
In recent years, fine-tuning on the downstream RE tasks with pre-trained models~\citep{soares2019matching} has achieved significant progress with the rapid development of the “Pre-train and Fine-tune” Paradigm~\citep{devlin2018bert} which leverages large-scale unlabeled data. However, RE still suffers from the data scarcity problem. For most RE tasks,  due to the task-specific definition of relations, the lack of customized annotation data poses great challenge for the supervised RE ~\citep{hendrickx2010semeval,zhang2017position}. Meanwhile, manually labeling large-scale RE data is extremely time-consuming, expensive, and laborious. 
As an alternative, automatically building annotated data for RE attracts a lot of attention in the research community~\citep{mintz_distant_2010,luo2019semi,yu-etal-2020-improving}.

Self-training is a simple and effective approach to build auto-annotated data~\citep{xie2020self,du2021self}.
The idea is to use a teacher model trained on human-annotated data to  annotate the additional unlabeled data. Then the human-annotated data is combined with some instances selected from the auto-annotated data to train a student model. In this paper, we follow the self-training framework to improve \textbf{Low-Resource RE}, which is closer to practical situations where the task starts with a small seed set of human-annotated data.

\begin{figure}[t]
    \centering
    \subfloat[Confident instance]{ \includegraphics[width=0.45\textwidth]{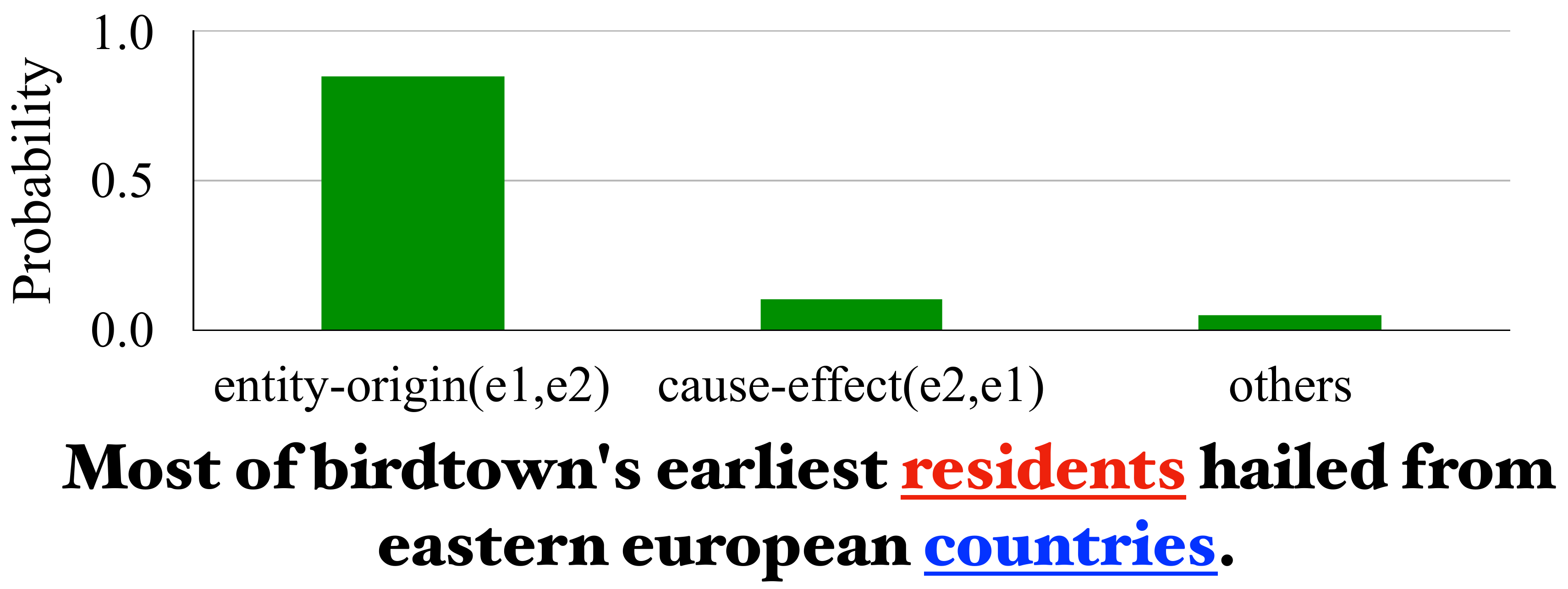}} \\
    \subfloat[Uncertain instance]{ \includegraphics[width=0.45\textwidth]{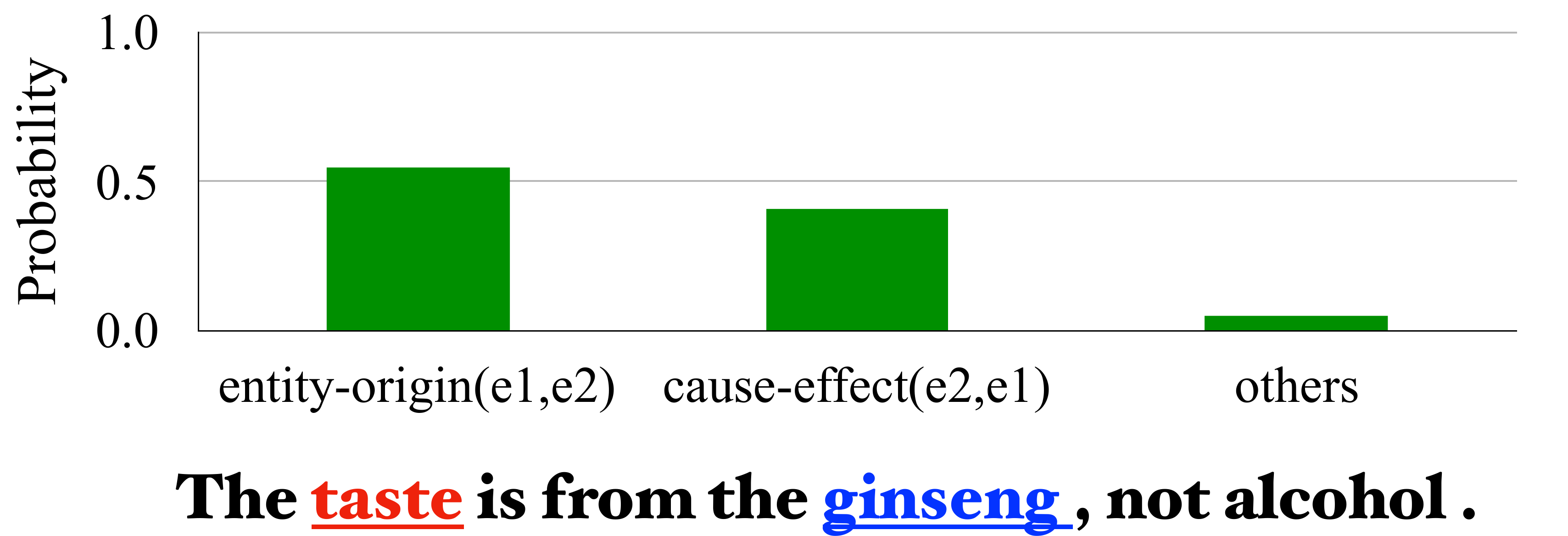}}
    \caption{Two examples of auto-annotated instances. For simplicity, we list two detailed relations and use  ``others'' to represent the other relations.
    }
    \label{fig:example}
\end{figure}

In previous studies, researchers often select the auto-annotated instances with high confidence, named as \emph{confident instance}, and have achieved a certain success~\citep{qian2009semi,du2021self}. Figure~\ref{fig:example}(a) shows an example of confident instance, where the teacher model can easily classify it as relation ``entity-origin(e1,e2)'' with the clue offered by ``hailed from''. Therefore, we first follow this kind of self-training solutions to train the RE system. However, in the preliminary experiments, we find that \emph{some relations might have similar expressions among instances}, which makes the teacher model confused.
As a result, for the uncertain instances, the teacher model gives similar high probabilities to some relations or assigns low probabilities to all the relations.
An example of uncertain instance is shown in Figure~\ref{fig:example}(b), where the teacher model predicts the instance as relation ``entity-origin(e1,e2)'' with a probability of 56\%, as ``cause-effect(e2,e1)'' (the ground truth label) with 42\%, and as other relations with only 2\%. It is hard to distinguish between the first two relations as the expression ``... is from ...'' is often used for both. In previous studies, the uncertain instances are often discarded due to the confusion. However, we argue that ignoring all the uncertain instances may not be appropriate since they may contain useful information. For example, it is a good clue that the answer is one of the first two relations with a probability 98\% for the instance in Figure ~\ref{fig:example}(b).

Ideally, we would wish to make full use of all the auto-annotated instances to improve the RE system. But it is very hard due to the confusion problem. Therefore, we split the uncertain instances into two groups: \emph{ambiguous set} and \emph{hard set}. The ambiguous set contains the instances for which the teacher model predicts similar high probabilities on some relations, while the hard set contains those that the teacher model assigns low probabilities to all the relations. In this paper, we focus on the ambiguous set and propose an approach to use the ambiguous instances and the confident instances to improve the system of low-resource RE.

In our approach, we tackle two main issues when exploiting the ambiguous instances: 1) how to identify the candidate labels of ambiguous instances; 2) how to train a new model with the ambiguous instances.
For the first issue, we adopt a probabilistic accumulation approach to obtain a set of relations, that is, a set of candidate labels containing the vast majority of probabilities, and then label ambiguous instances with the candidate labels in the set.
To deal with the second issue, we make an assumption:
\emph{For the ambiguous instances, the teacher model does not know which relation is the exact answer, but it does know that \#1) the answer is (with high probability) in the candidate-label set (likes the first two relations in Figure~\ref{fig:example}(b)) and \#2) the answer is not one of the relations which are with very low probabilities (likes ``others'' in Figure~\ref{fig:example}(b)).} 
Under this assumption, we treat the ambiguous instances as  partially-labeled instances~\citep{yan2020partial}, where the relations are divided into a candidate-label set and a negative-label set for each ambiguous instance. Then, we propose a novel set-negative training method to learn from the ambiguous instances by the negative-label set. In order to build the final relation extraction system, we further propose a joint training method which 
supports both positive and set-negative training.

Our main contributions are as follows:
\begin{itemize}
    \item We propose STAD, a self-training framework, which supports learning from ambiguous data.
    \item We propose a method to classify the auto-annotated instances generated from teacher model into confident set, ambiguous set and hard set. The ambiguous instances are then treated as partially-labeled, which can reduce the effect of confused expressions. To our best knowledge, it is the first time that partial labeling is used to tag the auto-annotated instances in RE.
    \item In order to exploit the auto-annotated instances properly, we propose set-negative training for the ambiguous instances and positive training for the confident instances. 
    The set-negative training method can utilize the information that the answer is not in the set of the relations which have very low probabilities predicted by the teacher model. 
    And we further propose a joint training method to combine positive and set-negative training methods to build the final RE system.
\end{itemize}

To verify the effectiveness of our approach, we conduct experiments on two widely used datasets SemEval2010 Task-8 and Re-TACRED with low-resource settings.
Experimental results show that our proposed approach significantly outperforms the conventional self-training system which only samples confident instances and other baseline systems.

\section{Our Approach}

\begin{figure}
    \centering
    \includegraphics[width=0.45\textwidth]{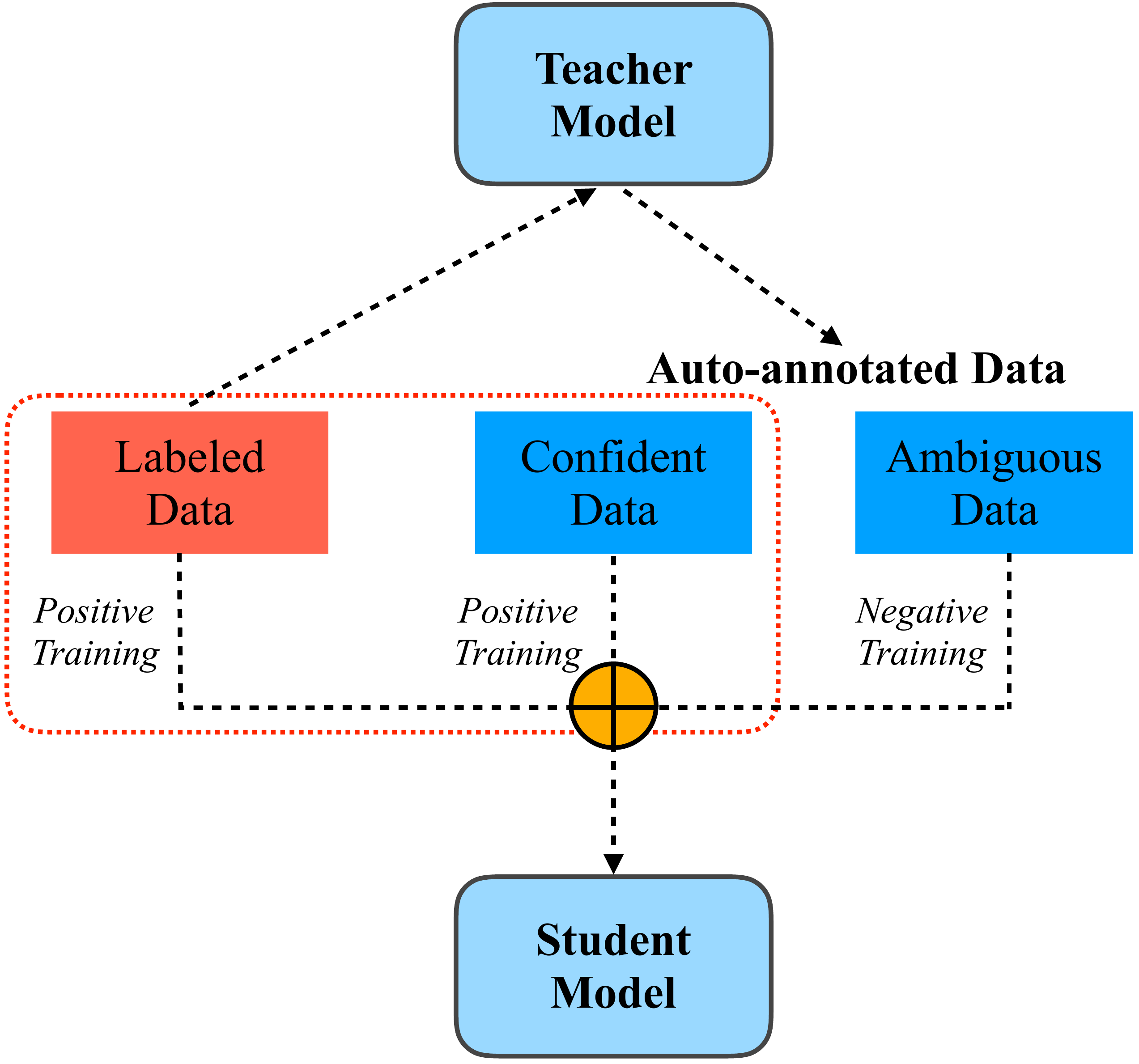}
    \caption{Framework of our approach. The red dotted rectangle is the conventional self-training approach that neglects the ambiguous data in auto-annotated data.}
    \label{fig:framework}
\end{figure}

We first briefly introduce the relation extraction task as well as the self-training framework frequently used in the previous studies~\citep{zhou2008semi}. Then, we propose an algorithm to classify the auto-annotated instances into three groups: confident data, ambiguous data and hard data.
Next, we introduce three tagging modes for ambiguous data.
Finally, we propose the training method to use both the confident and ambiguous data.
The framework of the proposed approach is shown in Figure~\ref{fig:framework}.

\subsection{Self-Training for Relation Extraction}
\label{self-training relation extraction}

\subsubsection{Relation Extraction}
\label{relation_extraction}

Fine-tuning on a pre-trained model, e.g., BERT, with a task-specific classifier is a common practice for downstream NLP tasks~\citep{devlin2018bert}.
Following~\citet{soares2019matching}, the RE model is composed of a BERT encoder and a relation classification layer. Entity markers (`[E1] head entity [/E1]' and `[E2] tail entity [/E2]') are inserted into input tokens to wrap the entities. Concretely, the output representation of two start entity markers (`[E1]' and `[E2]') are concatenated as the representation of entities. Finally, the representations are used as input for the relation classification layer.

Formally, the output representation of an input instance $x$ after BERT is $\mathbf{h} = \mathbf{h}_{[E1]} \oplus \mathbf{h}_{[E2]}$.
Then, the output probability distribution for $M$ relations $p=[p_1, p_2, ..., p_M]$ is computed by the relation classification layer:
\begin{equation}
    p = f(x) = \textrm{Softmax}(\mathbf{Wh}+\mathbf{b}),
    \label{encoder_eq}
\end{equation}
where $\mathbf{W}$ and $\mathbf{b}$ are model parameters.

During training, each instance $x$ from the human-annotated data is labeled with a one-hot label vector $y$: a single 1 value for the ground-truth label and 0 values for other labels.
Then, the positive training is performed to calculate the cross entropy loss:
\begin{equation}
    \label{eqn:positive train}
    \mathcal{L}_{PT}(f(x), y) = - \sum^{M}_{i=1}y_i \log{p_i},
\end{equation}
where $M$ is the number of relations, and $y_i$ and $p_i$ are the label and prediction probability of $i$th relation, respectively.

\subsubsection{Self-Training}
\label{self-training}
Generally, in the self-training framework, unlabeled instances are labeled by the teacher model to build the auto-annotated data. As shown in Figure~\ref{fig:framework}, the common-used flow of self-training is performed in the following steps: (1) use the human-annotated data to train a teacher model; (2) use the teacher model to conduct label prediction for unlabeled data; (3) select confident auto-annotated instances via a pre-defined probability threshold (described in Sec.~\ref{inst_classify}) and 
the rest are uncertain instances; (4) combine confident auto-annotated data and human-annotated data to train a student model (the red dotted rectangle in Figure~\ref{fig:framework}).

And in step (3), the remaining uncertain instances are considered to be useless. However, as described in Sec.~\ref{intro}, the uncertain instances (e.g., the example in Figure~\ref{fig:example}(b)) might contain useful information.

\subsection{Instance Classification}
\label{inst_classify}
As shown in Algorithm~\ref{alg:algorithm}, we propose a method to classify the auto-annotated data into confident data (Line 14), ambiguous data (Line 16) and hard data (Line 18).
In detail, considering a classification task with $M$ classes, we first set a probability threshold $T$.
After sorting the classes by the predicted probability from large to small, we dynamically accumulate the probability of top classes until the score is larger than $T$ (Line 6-10 in Algorithm~\ref{alg:algorithm}). Then, we move on to the classification of instances:

\paragraph{Confident instance.} Using a probability threshold to select confident instances is a common practice in self-training. We also add an instance into the confident data when the highest prediction probability exceeds the threshold $T$, as the teacher makes a certain prediction. In Line 13-14 of Algorithm~\ref{alg:algorithm}, instances with only one candidate label are added into the confident data $\mathbf{D_{con}}$. 
\paragraph{Ambiguous instance.} 
Besides the confident data, according to our observations, the teacher model may give relatively high probabilities to multiple relations.
In order to make full use of ambiguous instances to improve the RE system, we adopt a greedy probability accumulation method to identify ambiguous data from uncertain data. Specifically, the size of candidate-label set is at least two and can be at most $M-1$ (Line 15-16 in Algorithm~\ref{alg:algorithm}).
Obviously, as the size of candidate-label set increases, instances carry less information.

\paragraph{Hard instance.}
We also consider an extreme situation that the sum probability of top $M-1$ relations is still lower than $T$. In other words, the lowest predicted probability is larger than $1-T$. We treat such examples as hard instances because the teacher model is confused about all relations. 

\begin{algorithm}[tb]
\caption{Instance Classification}
\label{alg:algorithm}
\textbf{Input}: auto-annotated data $\mathbf{D_{auto}}=\{x, P, Y\}$ containing sentence $x$, prediction distribution $P$ and its corresponding relations $Y$.\\
\textbf{Hyper Parameter}: probability threshold $T$.\\
\textbf{Output}: confident data $\mathbf{D_{con}}$, ambiguous data $\mathbf{D_{amb}}$ and hard data $\mathbf{D_{hard}}$

\begin{algorithmic}[1] 
\FOR{$(x, P) \in \mathbf{D_{auto}}$}
    \STATE Let $score=0.0$.
    \STATE Let candidate-label set $C^+=\{\}$
    \STATE Sort $P$ from large to small
    \STATE Sort $Y$ by the order in $P$
    \FOR{$(p, y) \in {P, Y}$}
        \STATE $score \leftarrow score + p$.
        \STATE Append y to $C^+$.
        \IF {$score > T$}
            \STATE \textbf{break}
        \ENDIF
    \ENDFOR
    \IF {len($C^+$) == 1}
            \STATE Append $(x, C^+)$ to $\mathbf{D_{con}}$
    \ELSIF {len($C^+$) $\leq$ $M-1$}
            \STATE Append $(x, C^+)$ to $\mathbf{D_{amb}}$
    \ELSE
        \STATE Append $x$ to $\mathbf{D_{hard}}$
    \ENDIF
\ENDFOR
\STATE \textbf{return} $\mathbf{D_{con}}$, $\mathbf{D_{amb}}$, $\mathbf{D_{hard}}$
\end{algorithmic}
\end{algorithm}

\begin{table}[ht]
    \centering
    \setlength{\tabcolsep}{4pt}
    \begin{tabular}{lccc}
        \toprule
        \bf Mode & \bf Ent-Ori & \bf Cau-Eff & \bf Others \\
        \midrule
        Probability &  0.56 & 0.42 & 0.02 \\
        \hdashline
        ~~~~Hard Label & 1 & 0 & 0 \\
        ~~~~Soft Label & 0.56 & 0.42 & 0.02 \\
        ~~~~Partial Label & 1 & 1 & 0 \\
        \bottomrule
    \end{tabular}
    \caption{An example of three tagging modes with given predicted probability distribution. }
    \label{tab:tagging_mode}
\end{table}
\subsection{Instance Label Tagging Mode}
After identifying confident and ambiguous data from the auto-annotated data, we now have three training sets: a small seed set of human-annotated data $\mathbf{D_{hum}}$, a confident auto-annotated data $\mathbf{D_{con}}$, and an ambiguous auto-annotated data $\mathbf{D_{amb}}$.
For the  human-annotated data and confident data, we take the one-hot label vector as described in Sec.~\ref{relation_extraction} to tag the data. 
For the ambiguous data, a variety of methods can be used to tag the instance. As shown in Table~\ref{tab:tagging_mode}, given the probability distributions predicted by the teacher model, hard label mode assigns an exact label (the label with highest prediction probability) with a one-hot vector~\citep{lee2013pseudo} while soft label mode~\citep{xie2020self} adopts probability distributions as labels. 

In this work, we propose to use the partial label mode~\citep{yan2020partial} to tag ambiguous instances.
As the teacher model gives relatively high probabilities to relations in the candidate-label set $C^+$ (Line 3 in Algorithm~\ref{alg:algorithm}), partial label mode assigns each ambiguous instance with a set of candidate relations and treats each candidate relation equally to form a multi-hot label vector. As the example shown in Table~\ref{tab:tagging_mode}, the instance is labeled as [1, 1, 0] because the first two relations are in candidate-label set.

\subsection{Model Training}
As described in Sec.~\ref{relation_extraction}, we can directly utilize Eqn.~\ref{eqn:positive train} to train data tagged in the hard or soft label mode. 
In this section, we focus on training ambiguous data under partial label mode and describe how to train model on a mixed data.

\subsubsection{Set-Negative Training}
\label{NTPL}

Inspired by negative training~\citep{kim2019nlnl} which trains noisy data in a negative way by selecting a random label excepting the tagged label, we propose the set-negative training method to train ambiguous data.
For the ambiguous data, as described in Sec.~\ref{inst_classify}, we first split relations into two sets: candidate-label set $C^+$ and negative-label set $C^-$, where $C^-$ includes the relations that are not in $C^+$.
With the Assumption \#2 described in Sec.~\ref{intro}, we are confident that the answer is not in $C^-$. 
Hence, we randomly select one label from $C^-$ as a negative label for each ambiguous instance in every iteration to update the model in a negative way.

Formally, the loss function for the ambiguous instances under set-negative training is:

\begin{equation}
    \label{negative train}    \mathcal{L}_{NT}(f(x), y) = - \sum^{M}_{i=1}y_i \log{(1 - p_i)},
\end{equation}
where the one-hot label $y$ is dynamically changed by randomly selecting a negative label from $C^-$ for every iteration. 

\subsubsection{Joint Training}
\label{joint_train}

After adopting the set-negative training for ambiguous data, another problem is how to train on $\mathbf{D_{hum}}$, $\mathbf{D_{con}}$, and $\mathbf{D_{amb}}$ simultaneously. Formally, we need to design a unified training framework to support both positive training (Eqn.~\ref{eqn:positive train}) and set-negative training (Eqn.~\ref{negative train}). To this end, we first introduce a flag variable $z$ to represent whether current input instance is partially labeled or not:
\begin{equation}
    z = \left\{ \begin{array}{ll}
1 & \textrm{if partially labeled,}\\
0 & \textrm{others.}\\
\end{array} \right.
\end{equation}
Then, the following unified loss function can be directly used for joint training:
\begin{equation}
\label{unified_loss}
    \mathcal{L}(f(x), y) = - \sum^{M}_{i=1}y_i \log{|z - p_i|},
\end{equation}
where $|*|$ is the absolute value. When the input instance is partially labeled, this function is equivalent to Eqn.~\ref{negative train}, otherwise it is equivalent to Eqn.~\ref{eqn:positive train}.
\section{Experiments}
In this section, we carry out experiments to show the effectiveness of the proposed approach. We first introduce settings of datasets and parameters. Then, we describe comparison systems used in this work. After that, we present the overall evaluation results, followed by further analysis.

\subsection{Datasets}
We conduct our experiments on two widely used relation extraction datasets: SemEval 2010 Task-8 (SemEval) and Re-TACRED. The brief information of two datasets are as follows:
\begin{itemize}
    \item SemEval: A classical dataset in relation extraction which contains 10,717 annotated sentences covering 9 relations with two directions and one special relation ``no\_relation''~\citep{hendrickx2010semeval}.
    \item Re-TACRED: A repaired version of TACRED~\citep{zhang2017position} proposed by~\citet{stoica2021re} who re-annotated the data and refined relation definitions. In total, it contains 91,467 sentences covering 40 relations (also including a ``no\_relation'' class).
\end{itemize}

To test on the \textbf{low-resource scenario}, we randomly sample 20 instances for each relation from the original training set as the human-annotated data and use the remaining instances as unlabeled data.~\footnote{For any relation contains less than 20 instances in the original training set, we directly repeat some instances to keep the data balanced.} Meanwhile, we also rebuild the development set by sampling 10 instances for each relation from the original development set which is more realistic. 
Besides, the special ``no\_relation'' type in the Re-TACRED is excluded as it occupies more than 66\% of instances.
The statistics of two low-resource datasets are shown in Table~\ref{tab:data_statictic}.

\begin{table}[htb]
    \setlength{\tabcolsep}{3pt} 
    \centering
    \begin{tabular}{lrrrrr}
    \toprule
         \bf Data & \bf Rel & \bf Train & \bf Dev & \bf Test & \bf Unlabel \\
         \midrule
         SemEval & 19 & 380 & 190 & 2,717 & 6,076 \\
         Re-TACRED & 39 & 780 & 390  & 5,648 & 18,938 \\
         \bottomrule
    \end{tabular}
    \caption{Statistics of SemEval and Re-TACRED with the low-resource setting.}
    \label{tab:data_statictic}
\end{table}

\begin{table*}[]
    \centering
    \begin{tabular}{c l c c c c c }
    \toprule
       \multirow{2}{*}{\bf \#} & \multirow{2}{*}{\bf System} & \multicolumn{2}{c}{\bf Auto-annotated }  & \multicolumn{3}{c}{\bf Micro F1} \\
        &  & \bf $\mathbf{D_{con}}$ & \bf $\mathbf{D_{amb}}$ & \bf SemEval & \bf Re-TACRED  & \bf Avg. Score\\
         \midrule
    1     & \textsc{Supervised} & \texttimes & \texttimes & 77.3$_{\pm0.8}$ & 77.5$_{\pm1.3}$ & 77.4\\
        \midrule
    2     & \textsc{Self-Training} & \checkmark & \texttimes & 79.2$_{\pm1.5}$ & 80.5$_{\pm0.4}$ & 79.9\\
    3    & \textsc{Hard-Label}   & \checkmark& \checkmark & 78.0$_{\pm1.2}$ & 78.2$_{\pm1.4}$ & 78.1\\
    4    & \textsc{Soft-Label}   & \checkmark & \checkmark & 79.1$_{\pm0.8}$ & 79.8$_{\pm2.0}$ & 79.5\\
    5    & \textsc{STAD} (Ours.) & \checkmark & \checkmark & \bf 80.0$_{\pm1.0}$ & \bf 81.6$_{\pm0.3}$ & \bf 80.8\\
        \bottomrule
    \end{tabular}
    \caption{ Results on the test set of SemEval and Re-TACRED with low-resource settings in terms of micro F1 score. 
    $\mathbf{D_{con}}$ and  $\mathbf{D_{amb}}$ denote confident data and ambiguous data, respectively. We run experiment five times with different random seeds and report the mean of the micro F1 scores (Micro-F1) and its standard deviation score. Some additional experiments with random data samples are included in Appendix A.}
    \label{tab:main_results}
\end{table*}

\subsection{Parameter Settings}
\paragraph{Relation Extraction Model Training}
To train the relation extraction model, we use the ``Entity marker$+$Entity start state'' architecture proposed in~\citet{soares2019matching} with $\tt BERT_{base}$~\citep{devlin2018bert}.
During training, we follow~\citet{devlin2018bert} to select the learning rate among \{2e-5, 3e-5, 5e-5\}, batch size among \{16, 32\} and adopt the hyper parameters with the best performance on development set. For other parameters, we simply follow the settings used in~\citet{soares2019matching} to conduct experiments. We train the model in 20 epochs with early stop strategy to relieve the overfitting problem. Besides, we run each system five times with random seeds and report the mean of the micro F1 score (Micro-F1) with a standard deviation score. 

\paragraph{Auto-Annotated Data Building}
The probability threshold is an important hyper parameter which is used to select confident data and determine the candidate-label sets of ambiguous data.
In detail, we search probability threshold $T$ among \{0.95, 0.90, 0.85, 0.80\} on development set to select the best confident data. 

\subsection{Comparison Systems}
In order to make a fair comparison, all systems in this work use the BERT-based fine-tuning model proposed by~\citet{soares2019matching} as relation extraction model. Based on the difference of data parts and tagging strategies, we conduct following models for comparison.

\paragraph{\textsc{Supervised:}} A supervised baseline system for our relation extraction task~\cite{soares2019matching}. This system is trained only on human-annotated data. We also use this system as the teacher model to tag unlabeled data for the following approaches.
\paragraph{\textsc{Self-Training:}} A common-used self-training method~\cite{du2021self}, which combines human-annotated and confident data. We re-produce this self-training framework with the same relation extraction model of \textsc{Supervised} system.


\paragraph{\textsc{Hard-Label:}}
A direct comparison of the system proposed by~\citet{lee2013pseudo}, which extends the \textsc{Self-Training} system to involve all auto-annotated data.
We obtain the class with the highest prediction probability as the label for all auto-annotated data.
\paragraph{\textsc{Soft-Label:}}
Another extension of \textsc{Self-Training} system which utilizes the remaining auto-annotated data by assigning probability distributions comes from teacher model as labels~\citep{xie2020self}.

\subsection{Overall Evaluation Results}
\label{overall_results}


Table~\ref{tab:main_results} shows the overall evaluation results on SemEval and Re-TACRED with low-resource settings. Our observations are:
\begin{itemize}
\item System \textsc{Self-Training} significantly and consistently outperforms \textsc{Supervised} with +2.5 on Micro-F1 (79.9 vs. 77.4) in average. The results indicate that self-training with sampling confident data is effective for relation extraction in low-resource scenarios.

\item Systems \textsc{Hard-Label} and \textsc{Soft-Label} employ the ambiguous data outperform the \textsc{Supervised} system,  but can not achieve improvement over the \textsc{Self-training} system, demonstrating that it is challenging to achieve improvement with the ambiguous data. 

\item Our final system \textsc{STAD} significantly outperforms the \textsc{Supervised} system on both datasets with +3.4 (80.8 vs. 77.4) in average. And it also performs better than the \textsc{Self-Training} system, demonstrating the the effectiveness and the versatility of the proposed approach.  

\end{itemize}


\subsection{Extremely Low-resource Scenario}
In addition, we conduct additional experiments on \textbf{extremely low-resource scenario} with 15, 10 and 5 instances per relation to investigate the effect of the proposed approach. Experimental results are shown in Table~\ref{tab:low_sem} and Table~\ref{tab:low_rec}. 
From the two tables, we can find that:
\begin{itemize}
    \item The performance of \textsc{Supervised} drops rapidly when the amount of training data decreases. 
    \item \textsc{Self-Training} system works well on 20, 15 and 10, but it only achieves a slight improvement on 5. This indicates that it becomes hard to learn from confident data when teacher model is weak.
    \item The results of $\bigtriangleup_1$ and $\bigtriangleup_2$ show that our \textsc{STAD} system outperforms both \textsc{Supervised} and \textsc{Self-Training} on all settings, especially for the improvement obtained on data size is 5 which indicating the robustness of our system. We think the reason is that when the teacher model becomes weak, there is more and more ambiguous data in the auto-annotated data.
\end{itemize}

\begin{table}[t]
    \setlength{\tabcolsep}{5pt}
    \centering
    \begin{tabular}{c|rrrr}
    \toprule
        \multirow{2}{*}{\bf System} &  \multicolumn{4}{c}{\bf Data Size} \\
         & 20 & 15 & 10 & 5  \\
        \midrule
        \textsc{Supervised}        & 77.3 & 72.0 & 60.6  &  45.1    \\
        \textsc{Self-Training}    & 79.2 & 77.1 & 71.3 & 47.4   \\
        \midrule 
        \textsc{STAD}           & \bf 80.0 & \bf 78.1 & \bf 72.6 & \bf 53.5    \\
        $\bigtriangleup_1$  &  2.7 & 6.1  & 12.0  & 8.4\\
        $\bigtriangleup_2$  &  0.8 & 1.0  & 1.3  & 6.1\\
        \bottomrule
    \end{tabular}
    \caption{Results on the test set of SemEval with low-resource settings in terms of Micro F1 score. $\bigtriangleup_1$ and $\bigtriangleup_2$ represent the absolute improvement when comparing ours with supervised baseline and self-training baseline, respectively. }
    \label{tab:low_sem}
\end{table}

\begin{table}[t]
    \setlength{\tabcolsep}{5pt}
    \centering
    \begin{tabular}{c|rrrr}
    \toprule
         \multirow{2}{*}{\bf System} &  \multicolumn{4}{c}{\bf Data Size} \\
         & 20 & 15 & 10 & 5 \\
        \midrule
        \textsc{Supervised}        & 77.5 & 74.1 & 62.0 & 47.0     \\
        \textsc{Self-Training}     & 80.5 &   79.3   & 68.4 & 50.1     \\
        \midrule 
        \textsc{STAD}            & \bf 81.6 & \bf 81.1      & \bf 72.3 & \bf 60.0 \\
        $\bigtriangleup_1$ &  4.1 &  7.0   & 10.3  & 13.0 \\
        $\bigtriangleup_2$ &  1.1 &  1.8   & 3.9  & 9.9\\
        \bottomrule
    \end{tabular}
    \caption{Results on the test set of Re-TACRED with low-resource settings in terms of Micro F1 score. $\bigtriangleup_1$ and $\bigtriangleup_2$ represent the absolute improvement when comparing ours with supervised baseline and self-training baseline, respectively. }
    \label{tab:low_rec}
\end{table}

\subsection{Ablation Study}
\label{ablation_study}
\begin{table}[htb]
    \centering
    \begin{tabular}{lr}
    \toprule
         \bf Method & {\bf Mirco F1} \\
        \textsc{STAD} & 81.6  \\
        ~~~ $-$ Partial Labeling & -1.4 \\
        ~~~ $-$ Set-Negative Training & -2.3\\
        ~~~ $-$ Both & -3.4 \\
        \bottomrule
    \end{tabular}
    \caption{An ablation study for using partial labeling and set-negative training on the test set of Re-TACRED.}
    \label{tab:ablation}
\end{table}

To further analyze the effect of partial labeling and set-negative training on the ambiguous data in our proposed approach, we perform ablation studies on the low-resource Re-TACRED dataset and list the results in Table~\ref{tab:ablation}. 

Without the partial labeling, we label the ambiguous data in the hard label mode and use the negative training to train the model, the performance of our proposed approach degrades by 1.4 F1 score.  Without the set-negative training, we label the ambiguous data in partial label mode but with positive training. The strategy results in a drastic drop in F1 score. Removing both the partial labeling and set-negative training, the proposed system deals the ambiguous data in the same way as in  \textsc{Hard-Label} system in Table~\ref{tab:main_results} and suffers from performance degradation (78.2-81.6=-3.4).

These results demonstrate that the partial labeling and the set-negative training on the ambiguous data indeed contribute to the final performance.

\subsection{Distribution of Auto-Annotated Data}
\begin{figure}
    \centering
    \includegraphics[width=0.48\textwidth]{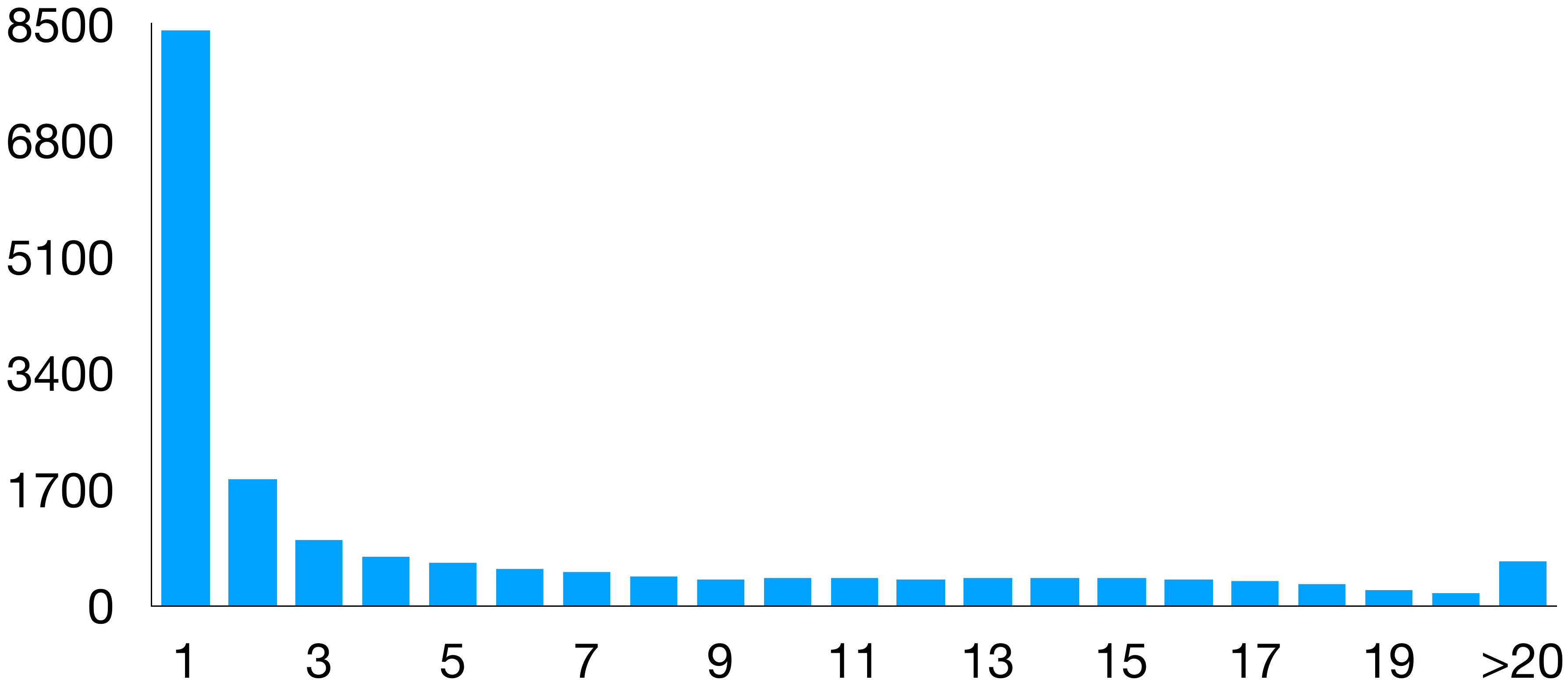}
    \caption{The number of auto-annotated instances according to size of candidate-label sets on Re-TACRED.}
    \label{fig:distribution}
\end{figure}

Figure~\ref{fig:distribution} shows the distribution of auto-annotated data under our probability accumulation method on Re-TACRED. From the figure we can find that although the number of confident data (the first bar, 44.5\%) is much larger than the others, the sum of ambiguous data (others, 55.5\%) is also considerable.
This indicates that ignoring such a large number of ambiguous data is not appropriate and how to make full use of them is the key to improving self-training.
In addition, we find that there is no hard instances in our experiments due to the large number of relation types and the setting of probability threshold.

\subsection{Top-N Evaluation}
\begin{figure}
    \centering
    \includegraphics[width=0.48\textwidth]{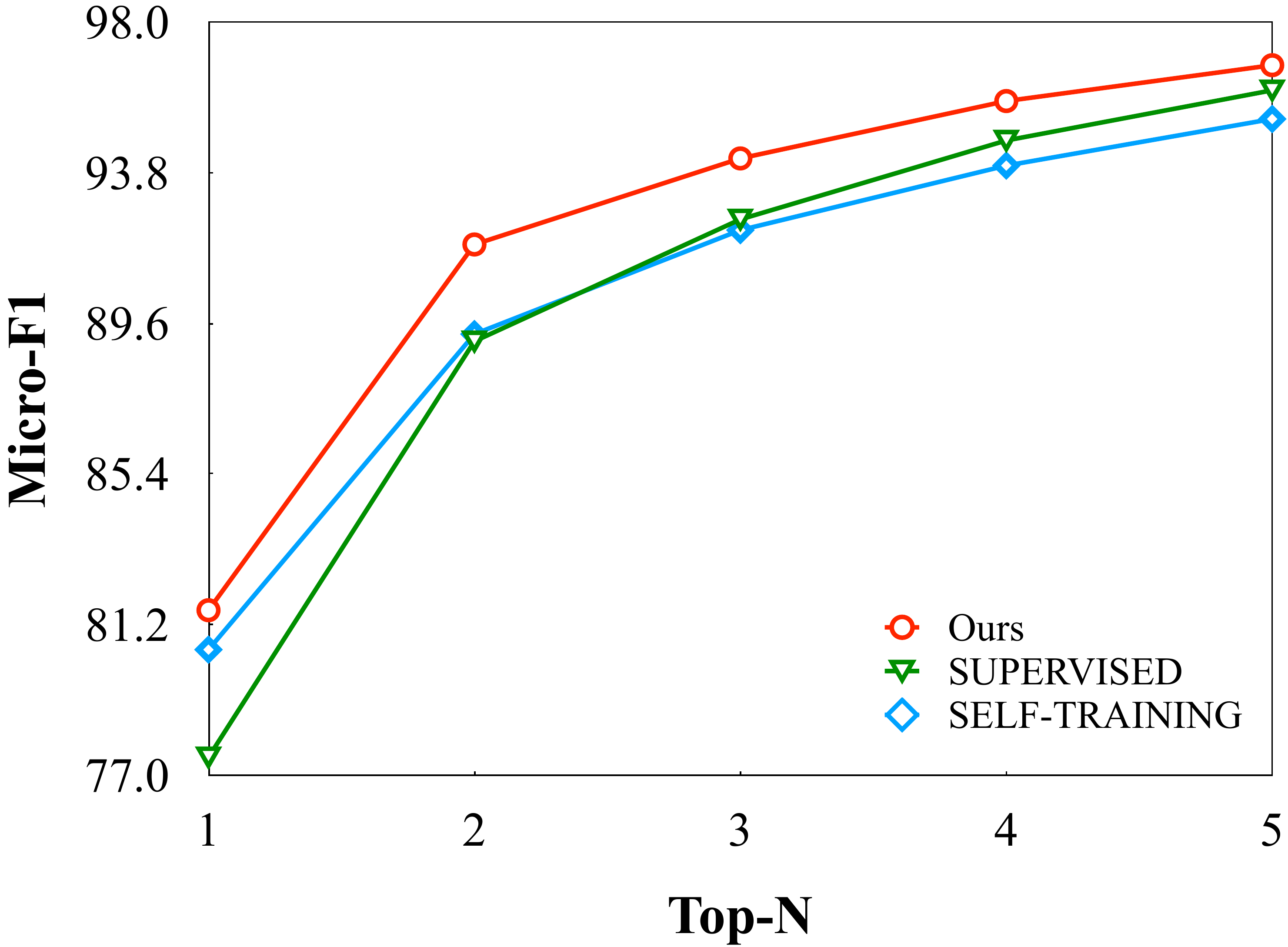}
    \caption{The top-N evaluation on Re-TACRED.}
    \label{fig:topn evaluation}
\end{figure}

Our method of modeling ambiguous data under negative training helps the model compress the scope of answers.
Intuitively, there should be more significant improvements in top-N evaluation. Therefore, we evaluate top-n predictions on Re-TACRED. The results are presented in Figure~\ref{fig:topn evaluation}. The figure shows that the \textsc{Self-Training} method shows a disadvantage in top-N evaluation as its performance gradually degrades when compared to the \textsc{Supervised} baseline. After modeling ambiguous data in a partially annotated and negative training mode, however, our method achieves significant and consistent improvement. In conclusion, our method of modeling ambiguous data can learn the information contained in the ambiguous data.

\begin{table}[ht]
    \centering
    \setlength{\tabcolsep}{3pt} 
    \begin{tabular}{l|ccc}
    \toprule
        \multirow{2}{*}{System} & \multicolumn{3}{c}{Data Sampling}  \\
        & Data 1 & Data 2 & Data 3 \\
    \midrule
        \textsc{Supervised} & 80.3$_{\pm1.6}$ & 79.7$_{\pm1.3}$ & 79.0$_{\pm1.5}$  \\
        \textsc{Self-Training} & 82.5$_{\pm0.6}$ & 84.0$_{\pm1.5}$ & 81.4$_{\pm1.7}$\\
        \textsc{STAD} & \bf 84.6$_{\pm1.5}$ & \bf 85.3$_{\pm1.0}$ & \bf 83.9$_{\pm1.6}$ \\
        \bottomrule
    \end{tabular}
    \caption{Results on Re-TACRED with different data samples for training and validation sets on the low-resource setting.}
    \label{tab:my_label}
\end{table}

\begin{table*}[h]
    \setlength{\tabcolsep}{3pt}
    \centering
    \begin{tabular}{c|c|c}
    \toprule
        Sentence & Relation &  Probability \\
        \midrule
         \multirow{2}{*}{\small It 's website , lists [Abu Zubaydah]$_{e1}$ 's birthplace as [Riyadh]$_{e2}$ , Saudi Arabia} & per:country\_of\_birth & 0.55\\
        & per:city\_of\_birth & 0.40 \\
        \midrule
         \multirow{2}{*}{\small [Parliament]$_{e2}$ speaker [Ali Larijani]$_{e1}$ , Iran 's former chief nuclear negotiator.}  & per:title & 0.79 \\
        & per:employee\_of & 0.16  \\
        \midrule
        \multirow{2}{*}{\small [Paul Kim]$_{e1}$ ( 25 ) Currently lives in [Saratoga]$_{e2}$ , CA ....} & per:city\_of\_birth & 0.74 \\
        & per:city\_of\_residence & 0.20 \\
       
        \bottomrule
    \end{tabular}
    \caption{Examples of ambiguous data in Re-TACRED.``e1'' and ``e2'' are head entity and tail entity respectively. ``Probability'' is predicted probability given by the teacher model. }
    \label{tab:example}
\end{table*}

\subsection{Experiments with Random Data Samples on Re-TACRED}
\label{sec:appendix}

In order to further study the stability of our approach, we conduct extra experiments on Re-TACRED which contains much more data than SemEval. For each experiment, we randomly sample a new split of training and validation sets for the low-resource setting. 
We report the mean of the micro F1 scores and its standard deviation score of five runs with different random seeds.

\subsection{Qualitative Results: Case Study}
\label{case_study}
Table~\ref{tab:example} shows examples of ambiguous data in Re-TACRED which are predicated by the teacher model.
We present the sentence as well as its relations with two highest predicted probabilities. The teacher model gives relatively high probabilities to the two relations and fails to predicate the ground-truth relation. 
For example, the teacher model can not distinguish ``per:city\_of\_birth'' from ``per:country\_of\_birth'' for the first sentence, and the conventional self-training system does not adopt this instance to the training set. But in our approach, this instance is tagged in a partially labeled mode and is used as an ambiguous instance to train the student model.

\section{Related Work}

\paragraph{Self-Training.}
Self-training is one of the most commonly used approaches for exploiting unlabeled data~\citep{scudder1965probability,yarowsky1995unsupervised,mcclosky2006effective,lee2013pseudo}. With the development of neural network models and the growth of demand for labeled data, self-training has become a very active field in research.
It is widely used to improve the performance of neural machine translation~\citep{zhang2016exploiting,jiao2020data,jiao2021self}, question answering~\citep{sachan2018self} and low resource dependency parsing~\citep{rotman2019deep}.
In this work, we apply self-training to exploit unlabeled data for low-resource relation extraction. The main difference is our work focuses on the uncertain instances which are often neglected in the previous studies.

\paragraph{Low-Resource Relation Extraction}
Recently, low-resource relation extraction (LRE) has attracted much attention due to data scarcity problems. There are two main scenarios of LRE: low-shot RE and low-resource supervised RE. Low-shot RE, including few-shot RE~\citep{han2018fewrel} and zero-shot RE~\citep{levy2017zero}, trains a model to identify unseen relation labels in a test set with a few support examples or even none of the examples. The low-resource supervised RE is designed to train a supervised model with small training data~\citep{deng2022knowledge}. In this work, we also focus on the low-resource supervised RE.

To solve the low-resource problem, one direction is to build a robust model which can make full use of small data and existing knowledge. For example, integrating semantic relation labels~\citep{dong2021mapre} and introducing knowledge graph~\citep{zhang2019long}. Another direction is to enlarge annotated data automatically. Data augmentation is a widely used technology to enrich the data~\citep{xu2016improved,yu-etal-2020-improving}. Self-training is also a conventional way to obtain automatically annotated data~\citep{rosenberg2005semi,hu2021semi,hu2021gradient}.

\paragraph{Partial Label Learning.}
In this work, the definition of partial label is a candidate set of labels for an ambiguous instance in a multi-class classification task~\citep{cour2011learning}.
This is different from that in sequence labeling tasks~\citep{li2014ambiguity} and multi-label multi-class classification tasks~\citep{xie2018partial}.
In order to learn from partially labeled instances,  various methods have been proposed to deal with the problem~\citep{nguyen2008classification,cour2011learning}. Recently, with the help of self-training, \citet{feng2019partial} proposes a self-guided retraining method to learn from partially labeled data.
Besides, \citet{yan2020partial} also proposes to recalculate the confidence of labels in a candidate set by taking the current model as a teacher.

\section{Conclusion}
This paper proposes a novel self-training approach with ambiguous data (STAD) for the low-resource relation extraction, which fully uses the auto-annotated data.  According to the probabilities predicted by the teacher model, we classify the auto-annotated data into three sets: confident set, ambiguous set and hard set. During training, we consider the ambiguous set, which is neglected by the previous studies, and adopt the set-negative training method to alleviate the noise problem.  Experimental results show that our proposed system consistently outperforms the 
self-training systems. 


\section*{Acknowledgements}
This work was supported by the National Natural Science
Foundation of China (61876115 and Grant No. 61936010),
the Priority Academic Program Development of Jiangsu
Higher Education Institutions. The corresponding authors are Wenliang Chen and Jiangjiang Zhao. We thank the anonymous reviewers for their constructive comments.
\bibliography{_main}
\bibliographystyle{acl_natbib}

\end{document}